\documentclass[sigconf]{acmart}
\AtBeginDocument{%
  \providecommand\BibTeX{{%
    \normalfont B\kern-0.5em{\scshape i\kern-0.25em b}\kern-0.8em\TeX}}}

\setcopyright{acmcopyright}
\copyrightyear{2023}
\acmYear{2023}
\acmDOI{XXXXXXX.XXXXXXX}

\acmConference[MM '23]{Make sure to enter the correct
  conference title from your rights confirmation emai}{October 29--November 3,
  2023}{Ottawa, Canada}
\acmPrice{15.00}
\acmISBN{978-1-4503-XXXX-X/18/06}

\begin{document}

\title{Hierarchical Semantic Perceptual Listener Head Video Generation: A High-performance Pipeline}

\author{Zhigang Chang}
\authornote{Corresponding Author}
\affiliation{%
  \institution{Du Xiaoman Financial}
  \institution{Shanghai Jiao Tong University}
  \city{Beijing-Shanghai}
  \country{China}
}
\email{changzhigang@duxiaoman.com}
\email{changzig@sjtu.edu.cn}

\author{Weitai Hu}
\affiliation{%
  \institution{Du Xiaoman Financial}
  \city{Beijing}
  \country{China}
}
\email{huweitai@duxiaoman.com}

\author{Qing Yang}
\affiliation{%
  \institution{Du Xiaoman Financial}
  \city{Beijing}
  \country{China}
}
\email{yangqing@duxiaoman.com}

\author{Shibao Zheng}
\affiliation{%
 \institution{Shanghai Jiao Tong University}
 \city{Shanghai}
 \country{China}}
\email{sbzh@sjtu.edu.cn}

\renewcommand{\shortauthors}{Trovato and Tobin, et al.}

\begin{abstract}
  In dyadic speaker-listener interactions, the listener's head reactions along with the speaker’s head movements, constitute an important non-verbal semantic expression together. The listener Head generation task aims to synthesize responsive listener's head videos based on audios of the speaker and reference images of the listener. Compared to the Talking-head generation, it is more challenging to capture the correlation clues from the speaker’s audio and visual information. Following the ViCo baseline scheme, we propose a high-performance solution by enhancing the hierarchical semantic extraction capability of the audio encoder module and improving the decoder part, renderer and post-processing modules. Our solution gets the first place on the official leaderboard for the track of listening head generation. This paper is a technical report of ViCo@2023 Conversational Head Generation Challenge in ACM Multimedia 2023 conference.  
  \end{abstract}




\maketitle

\section{Introduction}
In interactive conversational scenarios, facial expressions, head movements and micro-expressive reactions of both the speaker and the listener are capable of conveying important non-verbal semantics. Humans are able to acutely capture these non-verbal semantics while communicating verbally and respond in the same way. In dyadic speaker-listener interactions, we similarly expect machines to have the ability to perceive the speaker’s verbal and non-verbal semantics in multiple dimensions and simulate and generate the listener’s head response. This task is called “Responsive Listener Head Generation” which has great potential for applications in areas such as human-computer interaction, sentiment analysis, animation and metaverse. 

Compared to the Talking-head generation~\cite{wang2020mead,zhu2021deep}, there is much less research on listener head generation. This is because the mapping relationship between speech and speaker’s facial expressions and lip movements can be learned more strictly, while the listener’s reaction to the speaker’s audio and visual input is relatively ambiguous, that is, different head reactions can be appropriate and reasonable. However, when carefully comparing the videos of the speaker and the listener in an interaction, we can still find that the non-verbal semantic behaviors of the listener (such as nodding, smiling and head shaking) are feedback on the hierarchical multi-modal semantics of the speaker (such as linguistic emotions, speech rhythms, expressions and eyes contact), and carry personal behavioral styles. So the question now is how to extract the correlation clues between the speaker’s audio and video information and the listener’s non-verbal semantic behaviors.

Inspired by \cite{liu2022learning}, we adopt a Hierarchical Audio Encoder to fully exploit the semantics of different levels. Through this encoder, we believe that both high-level speech semantics (such as the emotions and tone conveyed by the content) and low-level speech semantics (such as the rhythm and pitch of the speech) can be captured and matched to the listener’s responsive behaviors. Similar to the ViCo baseline~\cite{zhou2022responsive}, we also use the face reconstruction model~\cite{guo2020towards} to extract the speaker's three-dimensional morphable model (3DMM) coefficients~\cite{blanz1999morphable, paysan20093d} as a visual signal encoder. A cascaded bi-directional GRU network acts as a decoder to predict the sequential outputs of listener head coefficients. In the final rendering and post-processing stage, we employ an enhanced renderer~\cite{tang2022explicitly} and a video restoration module~\cite{liang2022recurrent} to enhance the quality of the generated videos.

\begin{figure*}[htbp]
\vskip -0.5cm
\includegraphics[width=\textwidth]{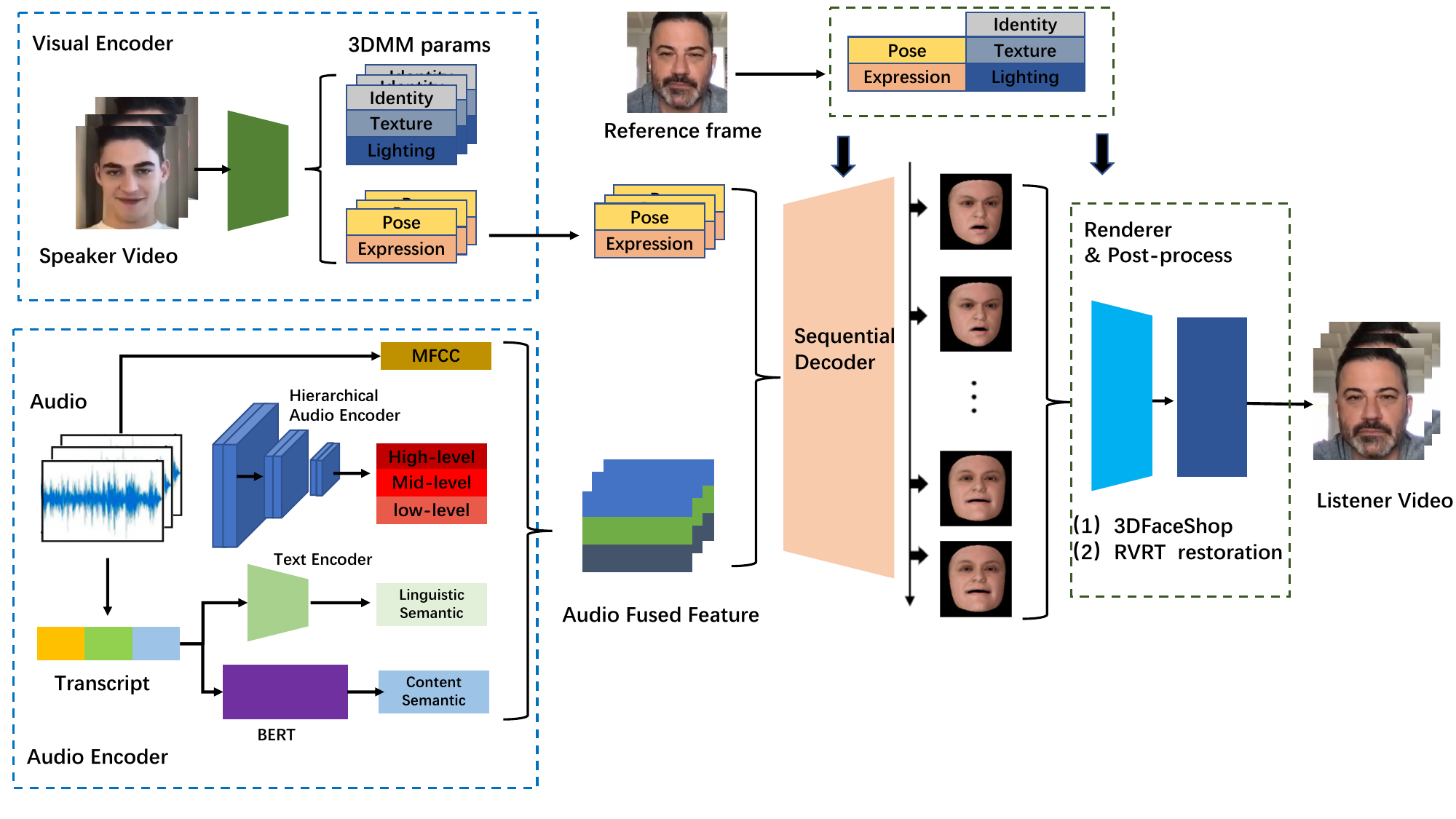}
\caption{The Pipeline of Hierarchical Semantic Perceptual Listener Head Video
Generation. }
\label{fig1}
\vskip -0.2cm
\end{figure*}

In general, this paper reports a high-performance solution to the Responsive Listener Head Generation task. Our main techniques include:
\begin{itemize}
    \item We adopt a Hierarchical Audio Encoder to capture hierarchical speech semantics, and design a audio feature fusion module to fuse audio features.
    \item We adopt a cascaded bi-directional GRU as a decoder to predict the sequential outputs. Several training techniques are applied on the limited training data.
    \item We adopt an enhanced renderer and a video restoration module to improve the quality of the generated videos significantly.
\end{itemize}

\section{Related Work}
Audio-Driven human-centric synthesis tasks~\cite{prajwal2020lip,wang2020mead,zhu2021deep,ginosar2019learning,liu2022learning} have attracted extensive research interest in recent years. The Audio-to-lip task~\cite{prajwal2020lip} aims to generate a lip-synchronization video clip with a initial frame input. The Talking-head generation task~\cite{wang2020mead,zhu2021deep} tends to output a vivid talking video of the speaker with head movements from a speaker's initial frame and a clip of audio. The Audio-to-gesture task~\cite{ginosar2019learning,liu2022learning} tries to learn mappings between audios and body languages. The commonality of the above generation tasks are that they are all based on the role of the speaker, and they all need to build the associations between speech semantics and target behaviors. Listening head generation task~\cite{zhou2023interactive, zhou2022responsive, luo2023reactface} is a non-verbal facial reaction prediction task. Recent facial reaction
studies~\cite{ng2022learning, zhou2022responsive} frequently represent facial motions with 3DMM coefficients (e.g., expression and pose coefficients) to better visualize the facial muscle movements. Meanwhile, speaker’s speech and audio behaviours were also additionally employed as the input to deliver richer verbal and non-verbal speaker behavioural cues~\cite{ng2022learning, song2022learning, shao2021personality}.

\section{Task Overview}
\subsection{Definition} 
We first briefly introduce the task definition of Responsive Listening Head Generation. Given an input video sequence $\mathcal{V}_{t}^{s}=\left \{ v_{1}^{s},\dots ,v_{t}^{s} \right \} $ of the speaker head in time stamps ranging from $\left \{ 1,\dots ,t \right \} $, and the corresponding audio signal sequence $\mathcal{A}_{t}^{s}=\left \{ a_{1},\dots ,a_{t} \right \} $, the goal of this task is to generate the whole listener's head video sequence $\mathcal{V}_{t}^{l}=\left \{ v_{2}^{l},\dots ,v_{t+1}^{} \right \} $, where output of the next time stamp can be denoted as Eq.~\ref{overall0} and $\mathcal{V}_{1}^{l}$ is the initial reference frame.
\begin{equation}
    \mathcal{V}_{t+1}^{l}=\mathbf{G} \left (\mathcal{V}_{t}^{s},\mathcal{A}_{t}^{s}, \mathcal{V}_{1}^{l} \right ) 
    \label{overall0}
\end{equation}

\subsection{Dataset Description and Evaluation Metrics}
This ViCo@2023 challenge is based on the extended ViCo dataset. The training set contains 440 pairs of speaker-listener videos with 83 listener IDs and 65 speaker IDs. The listening head test set contains 45 audio clips, speaker video clips and several listener reference images. 

The quality of generated videos will be quantitative evaluated from the prespectives of visual quality, naturalness, and task-specified goals. The evaluation meaning of each indicator is as follows:

\noindent The \textbf{PSNR} (Peak Signal to Noise Ratio) and \textbf{SSIM} (Structure Similarity Index
Measure) can evaluate the quality of generated images by comparing with GT images.

\noindent The \textbf{CPBD} (perceptual-based no-reference objective image sharpness metric) can evaluate the sharpness of the generated images without GT images.

\noindent The \textbf{FID} (Fréchet Inception Distance) evaluates the similarity of the generated images and GT in feature level.

\noindent The \textbf{CSIM} is evaluated by extracting identity features predicted by ArcFace and calculate the cosine similarity.

\noindent The \textbf{ExpL1} and \textbf{PoseL1} evaluate L1 distances of expression and pose coefficients from 3DMM reconstruction. 

\noindent The \textbf{ExpFD} and \textbf{PoseFD} evaluate Fréchet distances of expression and pose coefficients from 3DMM reconstruction. 

\begin{table*}[htbp]
\caption{The official final leaderboard for the track
of listening head generation in ViCo@2023 challenge (results of top-3 teams are shown here).}
\begin{center}
\label{table1}
\resizebox{0.95\textwidth}{!}{
\begin{tabular}{r|ccccccccc|c}
\hline
 Team  & SSIM $\uparrow$ & CPBD $\uparrow$ & PSNR $\uparrow$ & FID $\downarrow$ & CSIM $\uparrow$ & PoseL1 $\downarrow$ & ExpL1 $\downarrow$ &PoseFD $\downarrow$ & ExpFD $\downarrow$ & score $\downarrow$\\
 \hline\hline
\textbf{Ours} & \textbf{0.647} & 0.163 & \textbf{26.757} & 30.605 & 0.669 & \textbf{0.070} & \textbf{0.139} & \textbf{0.009} & 0.936 & 7\\
Robo Space & 0.612 & \textbf{0.186} & 19.064 & 30.982 & \textbf{0.676} & 0.078 & 0.149 & 0.024 & \textbf{0.604} & 9\\
ilearn & 0.597 & 0.168 & 18.398 & \textbf{29.134} & 0.636 & 0.085 & 0.154 & 0.012 & 0.618 & 15\\

\hline

\end{tabular}}
\end{center}
\end{table*}

\section{Method}

\subsection{Framework}
The overall pipeline is illustrated in Fig.~\ref{fig1}  The pipeline is composed of an audio encoder, a visual encoder, a sequential decoder, a video renderer and a post-processing module. Then we will introduce each component.  

\subsection{Visual Encoder}
Following the ViCo baseline scheme~\cite{zhou2022responsive}, 3DMM coefficients are extracted by a face reconstruction model~\cite{guo2020towards} for each frame. We get 3DMM coefficients $\mathcal{x}_{t}=\left (\mathcal{\alpha}_{t},\mathcal{\beta}_{t}, \mathcal{\epsilon}_{t}, \mathcal{\gamma }_{t},p_{t}   \right ) $ for each time stamp where the identity-independent features can be represented parametrically by expression and pose coefficients $\left \{ \mathcal{\beta}\in \mathbb{R}^{64},  p\in \mathbb{R}^{6} \right \}$. 

\subsection{Audio Encoder}
We adopt the Hierarchical Audio Encoder $E_{a}$ in HA2G~\cite{liu2022learning} to extract multi-level semantic features. Transcript can be directly recognized by Automatic Speech Recognition (ASR) models from audio$\mathcal{A}_{t}^{s}$. High-level linguistic features extracted by text encoder from transcripts $t_{text}$ as $f_{t}=E_{text}(t_{text})$. $E_{a}$ is a ResNetSE34 borrowed from \cite{chung2020defence}. Specifically, we define the features output from ResNet Stage-2 as low-level features $f_{a}^{low}$, features output from ResNet Stage-3 as middle-level audio features $f_{a}^{mid}$, features output from ResNet Stage-4 as high-level audio features $f_{a}^{high}$. We propose to learn the association between provided transcripts and audios with contrastive learning. The training strategy is to leverage the natural synchronization between text and audio. The multi-level contrastive loss are computed by Eq.~\ref{const_loss} to enforce the high-level audio features $f_{a}^{high}$ aligned with the text features $f_{t}$.
\begin{equation}
    L_{const}=-\log \frac{\exp \left ( sim\left ( f_{t},f_{a+}^{high} \right ) /\tau  \right ) }{ {\textstyle \sum_{i=1}^{K}\sum_{l\in L}\exp \left ( sim\left ( f_t,f_{a(i)}^{l} \right )/\tau   \right )  } } 
    \label{const_loss}
\end{equation}
where $L=\left \{ \mathrm{low},\mathrm{mid},\mathrm{high} \right \} $ and $sim(f_1,f_2)=\frac{f_1\cdot f_2}{\left | f_1 \right | \left | f_2 \right | }$ is cosine similarity. $f_{a+}^{high}$ is the high-level audio features aligned to the transcript as the positive samples. Negative samples are collected from high-level audio features of other time stamps or non-high-level semantic audio features. $\tau$ is the temperature parameter.

We also implement the pre-trained language model BERT~\cite{devlin2018bert} on transcripts to get the content semantics $S_{T} = \left \{ s_1,\dots ,s_t \right \}$. Mel-frequency cepstral coefficients (MFCC) features with the corresponding MFCC Delta and Delta-Delta (second-order
difference) features are extracted from the original speech as well, denoted as $M_{T} = \left \{ m_1,\dots ,m_t \right \} $.

An audio feature fusion module $f_{A-FFM}$ is designed to obtain a fusion of multiple speech features $f_{a}^{high}$, $f_{a}^{mid}$, $f_{a}^{low}$, $S_{T}$ and $M_{T}$. We adopt a SE-ResNet Module~\cite{hu2018squeeze} as the audio feature fusion module to capture the channel attention. The output of the audio encoder can be denoted as follows:
\begin{equation}
    A_{t}^{AE} =  f_{A-FFM}\left ( f_{at}^{high}, f_{at}^{mid}, f_{at}^{low}, S_{t}, M_{t} \right ) 
    \label{audio_en}
\end{equation}

\begin{table*}[htbp]
\caption{The quantitative results of several trial submissions on evaluation system. We select several results with different components as non-rigorous ablation experiments. More rigorous
ablation experiments should be re-organized. }
\begin{center}
\label{table2}
\resizebox{0.95\linewidth}{!}{
\begin{tabular}{r|ccccccccc}
\hline
 Settings  & SSIM $\uparrow$ & CPBD $\uparrow$ & PSNR $\uparrow$ & FID $\downarrow$ & CSIM $\uparrow$ & PoseL1 $\downarrow$ & ExpL1 $\downarrow$ &PoseFD $\downarrow$ & ExpFD $\downarrow$ \\
 \hline
 ViCo baseline & 0.495 &	0.152 &	15.335 & 41.319	& 0.538	 & 0.120 & 0.197 & 0.015 & 1.497\\
 Ours with PIRenderer & 0.612 &0.156	& 22.655 & 35.909 & 0.636 & 0.075 & 0.156 & 0.012 & 1.014\\  
 Ours with 3DFaceShop & 0.639 &0.161	&25.563	&32.290	& 0.639	& 0.072	&0.141	&0.010	&0.997	\\  
 Ours (full pipeline) & \textbf{0.647} & \textbf{0.163} & \textbf{26.757} & \textbf{30.605} & \textbf{0.669} & \textbf{0.070} & \textbf{0.139} & \textbf{0.009} & \textbf{0.936} \\  

\hline

\end{tabular}}
\end{center}
\end{table*}

\subsection{Sequential Decoder}
After acquiring audio and visual representations of the speaker, a cross-modal fusion function $f_{XM}$ is implemented to get the cross-modal comprehensive representations of the speaker:
\begin{equation}
    e_{t}^{xm} =  f_{XM}\left (\mathcal{\beta}_{t}, p_{t}, A_{t}^{AE}\right ) 
    \label{XM_en}
\end{equation}

At each time stamp $t$, the decoder module takes the he cross-modal comprehensive representation $e_{t}^{xm}$ as input and generate the listener's motion parameters $\mathcal{\beta}_{t}$ and $p_{t}$. The architecture of sequential decoder contains a six-layer bi-directional GRU. The decoding procedure can be formulated as:
\begin{equation}
    \mathcal{\beta}_{t+1}, p_{t+1}=\mathbf{G}_{\mathrm{GRU} }\left ( h_{t}, e_{t}^{xm} \right )  
    \label{gru}
\end{equation}
where $h_{t}$ is the $t$-th hidden state.

For the decoder model optimization, we supervise each prediction $\mathcal{\beta}_{t}, p_{t}$ with the groud truth (GT) of 3DMM coefficients in training set. The regression loss of the decoder is calculated as:
\begin{equation}
\begin{split}
    L_{dec} &=\sum_{t=2}^{T} || \beta_{t} - \hat{\beta_{t}} ||_{2} + {|| p_{t} - \hat{p_{t}}  ||}_{2} \\ &+\sum_{t=2}^{T}w_{1}{||\mu (\beta_{t})-\mu (\hat{\beta_{t}})||}_{2}+\sum_{t=2}^{T}w_{2}{||\mu (p_{t})-\mu (\hat{p_{t}})||}_{2}  
    \label{loss_reg}
\end{split}
\end{equation}
where $\hat{\beta_{t}}$, $\hat{p_{t}}$ denote the groud truth. The last two terms of Eq.~\ref{loss_reg} apply motion constraints to guarantee the inter-frame continuity, where $\mu \left ( \cdot  \right ) $ measures the inter-frame changes.

\subsection{Video Renderer and Post-processing Module}
The video renderer module in our pipeline refers to a controllable 3D-aware portrait generation model~\cite{ren2021pirenderer,tang2022explicitly}, which is able to convert the predicted sequential 3DMM coefficients (expressions and head poses) to video based on a reference image. The ViCo baseline adopts PIRenderer~\cite{ren2021pirenderer} as video renderer. Through observations, it can be found that PIRenderer performs well in general, but has several defects, including background distortion and poor performance on preserving the identity when large angles. Thus we have adopted a novel state-of-the-art renderer called 3DFaceShop~\cite{tang2022explicitly}. 3DFaceShop proposes a volume blending strategy where the renderer can form a composite output by blending dynamic and static areas, with two parts segmented from the jointly learned semantic field. With this blending strategy, 3DFaceShop could solve non-face area distortion and produce realistic portraits when viewed from free viewpoints. Experiments have demonstrated that the adoption of 3DFaceShop can lead to significant performance gains.

The final post-processing module is a video restoration method application, i.e., RVRT~\cite{liang2022recurrent}, to overcome the bottleneck due to the low-resolution and blurred training data. Specifically, we utilize pre-trained video restoration models to enhance facial details and improve visualization as much as possible.

\begin{figure}[htbp]
\vskip -0.5cm
\includegraphics[width=\linewidth]{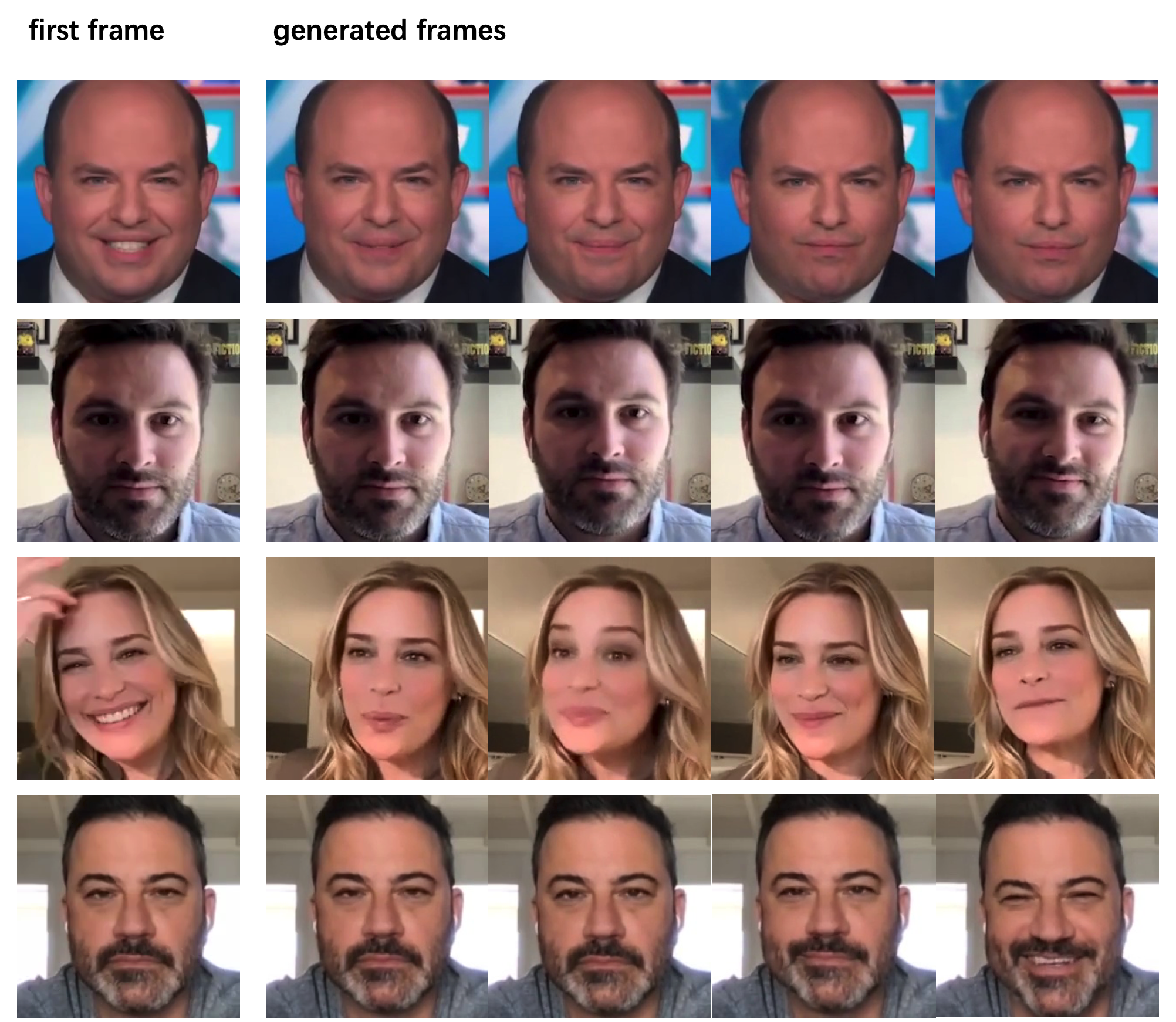}
\caption{Visualization results of generated listening head video frames based on first video frame.}
\label{fig2}
\vskip -0.7cm
\end{figure}

\section{Experiments}

\subsection{Overall Evaluation on the leaderboard}
As shown in Tab.~\ref{table1}, our proposed solution won five \textbf{1st places} on the official leaderboard out of nine evaluation metrics, and finally achieved the \textbf{1st place} in the overall score on the track of listening head generation. Our proposed solution leads the way in SSIM, PSNR, PoseL1, ExpL1 and PoseFD evaluation metrics, which indicates our generated videos are of excellent quality, both from visual prespective and from task-specified goals prespective.

\subsection{Limited Alation Study of submitted results}
In order to validate the effectiveness of the components of our proposed solution, we selected a few of the many results submitted to the online evaluation system for limited ablation experiments. Due to the large number of results submitted, the selected results do not guarantee that the experimental setups are aligned, and the results can only be used as a reference, with more rigorous ablation experiments to be re-organized. As shown in Tab.~\ref{table2}, our solution with PIRenderer has substantial performance gains compared to ViCo baseline, proving that our proposed audio encoder and sequential decoder are more capable of learning the associations between speech semantics and listener head behaviors. The experimental results also proved that the 3DFaceShop~\cite{tang2022explicitly} renderer was able to alleviate some of the problems that existed in the previous PIRenderer, bringing some performance gains.

\subsection{Qualitative Visualization}
We show the visualization results of our solution in Fig.~\ref{fig2}. Based on the reference image of the first frame in the leftmost column, it can be observed that our method is able to generate high-quality photo-realistic video clips of the listening heads.

\section{Conclusion}
This paper is a technical report of ViCo@2023 Conversational Head Generation Challenge in ACM Multimedia 2023 conference. Following the ViCo baseline scheme, we propose a high-performance solution by enhancing the herarchical semantic extraction capability of the audio encoder module and improving the decoder part, renderer and post-processing modules. Our solution gets the first place on the official leaderboard for the track of listening head generation.

\bibliographystyle{ACM-Reference-Format}
\bibliography{sample-base}

\appendix

\end{document}